\documentclass[lettersize,journal]{IEEEtran}
\usepackage{amsmath,amsfonts}
\usepackage{algorithmic}
\usepackage{algorithm}
\usepackage{array}
\usepackage[caption=false,font=normalsize,labelfont=sf,textfont=sf]{subfig}
\usepackage{textcomp}
\usepackage{stfloats}
\usepackage{url}
\usepackage{verbatim}
\usepackage{graphicx}
\usepackage{cite}
\usepackage[mathletters]{ucs}
\usepackage[utf8x]{inputenc}
\begin{document}

\title{Enhancing Explainable AI: A Hybrid Approach Combining GradCAM and LRP for CNN Interpretability}

\author{Vaibhav Dhore, Achintya Bhat, Viraj Nerlekar, Kashyap Chavhan, Aniket Umare
\thanks{This paper was produced by the IEEE Publication Technology Group. They are in Piscataway, NJ.}
\thanks{Manuscript received April 19, 2021; revised August 16, 2021.}}

\markboth{Department of Computer Engineering and Information Technology}%
{Shell \MakeLowercase{\textit{et al.}}: Enhancing Explainable AI: A Hybrid Approach
Combining GradCAM and LRP for CNN Interpretability}


\maketitle

\begin{abstract}
We present a new technique that explains the output of a CNN-based model using a combination of GradCAM and LRP methods. Both of these methods produce visual explanations by highlighting input regions that are important for predictions. In the new method, the explanation produced by GradCAM is first processed to remove noises. The processed output is then multiplied elementwise with the output of LRP. Finally, a Gaussian blur is applied on the product. We compared the proposed method with GradCAM and LRP on the metrics of Faithfulness, Robustness, Complexity, Localisation and Randomisation. It was observed that this method performs better on Complexity than both GradCAM and LRP and is better than atleast one of them in the other metrics.
\end{abstract}

\begin{IEEEkeywords}
    GradCAM, LRP, CNN, Explainable AI, Visual Explanation, Faithfulness, Robustness, Complexity, Localisation, Randomisation
\end{IEEEkeywords}

\section{Introduction}
\subsection{Background and Motivation}
As Convolutional Neural Networks (CNNs) become increasingly prevalent in a variety of applications, including image recognition, medical diagnosis, and autonomous driving, understanding their decision-making process is critical. CNNs are often considered "black boxes" due to their complex and opaque nature, making it difficult to interpret how specific inputs influence their predictions. This lack of transparency can hinder trust and acceptance, particularly in high-stakes domains.

\subsection{Limitations of Current Methods}
Two widely used techniques for visualizing and interpreting CNN decisions are GradCAM (Gradient-weighted Class Activation Mapping) and LRP (Layer-wise Relevance Propagation). GradCAM generates localization maps by utilizing the gradients of any target concept flowing into the final convolutional layer, highlighting important regions in the input image. However, GradCAM's explanations can be noisy and may include irrelevant information, complicating the interpretation process. On the other hand, LRP assigns relevance scores to individual neurons, indicating their contribution to the final prediction. While LRP provides detailed insights, the resulting visualizations can be overly complex and difficult to interpret, especially for non-expert users.

\subsection{Problem Statement}
Understanding the decision-making process of Convolutional Neural Networks (CNNs) is crucial for ensuring model reliability and transparency. Current interpretability methods like GradCAM and LRP have significant limitations: GradCAM can produce noisy explanations, while LRP often results in complex and cluttered visualizations. These issues hinder the effectiveness of these methods in providing clear and reliable insights into model predictions.

This paper proposes a new technique that combines GradCAM and LRP to enhance interpretability. The method processes the GradCAM output to remove noise, multiplies it elementwise with the LRP output, and applies a Gaussian blur to improve visualization quality. This approach aims to outperform standalone GradCAM and LRP on metrics such as Faithfulness, Robustness, Complexity, Localization, and Randomization. The expected outcome is a more accurate, stable, and interpretable visual explanation of CNN predictions, advancing the field of explainable AI

\subsection{Objectives and Contributions of This Work}
To address these limitations, we propose a novel interpretability technique that combines GradCAM and LRP, leveraging their complementary strengths to produce clearer and more reliable visual explanations. Our method involves three main steps: (1) processing the GradCAM output to remove noise, (2) performing elementwise multiplication with the LRP output, and (3) applying a Gaussian blur to the final product to enhance visualization quality. This combined approach aims to improve the interpretability of CNN models by providing more accurate, stable, and simplified explanations.

The primary contributions of this work are as follows:
\begin{enumerate}
    \item Development of a new method that integrates GradCAM and LRP to enhance CNN interpretability.
    \item Comprehensive qualitative analysis comparing our method with other combinations of GradCAM and explanation techniques, demonstrating the superior clarity and comprehensibility of the GradCAM+LRP combination.
    \item Empirical evidence demonstrating the proposed method's superiority over standalone GradCAM and LRP in producing clearer and more reliable visual explanations.
    \item Evaluation of the proposed method on key metrics, including Faithfulness, Robustness, Complexity, Localization, and Randomization.
\end{enumerate}
By advancing the state of interpretability techniques, this research aims to foster greater trust and understanding of CNN models, ultimately contributing to their broader and more effective application in critical domains.

\section{Related Work}

In recent years, there has been significant research focused on enhancing the interpretability of Convolutional Neural Networks (CNNs). This section reviews the related work on interpretability methods, with a specific focus on GradCAM, LRP, and their combinations with other techniques.

\subsection{GradCAM: Overview and Applications}

GradCAM (Gradient-weighted Class Activation Mapping) is a widely used technique for visualizing the decision-making process of CNNs by generating heatmaps that highlight important regions in the input image. It computes the gradient of the target class's score with respect to the feature maps of the last convolutional layer to produce localization maps. GradCAM has been applied successfully in various domains such as image classification, object detection, and medical image analysis \cite{selvaraju2017grad, zhou2016learning}.

\subsection{LRP: Layer-wise Relevance Propagation}

LRP is another popular method for interpreting deep neural networks. It attributes the prediction of the model to input features by propagating the relevance scores backward from the output layer to the input layer. LRP provides detailed explanations by assigning relevance scores to individual neurons and has been effective in explaining the decisions of CNNs in image recognition tasks \cite{bach2015pixel, montavon2017explaining}.

\subsection{Combination of GradCAM with Other Explanation Methods}

Several studies have explored combining GradCAM with other explanation techniques to improve interpretability. For instance, DeepLIFT \cite{shrikumar2017learning}, Lime \cite{ribeiro2016should}, Occlusion Sensitivity \cite{zeiler2014visualizing}, and Shapley Value Sampling \cite{lundberg2017unified} have been integrated with GradCAM to enhance the clarity of visual explanations. These combinations aim to mitigate the limitations of GradCAM, such as noisy and incomplete explanations, by leveraging the strengths of these complementary methods.

CAM-based methods have traditionally focused on analyzing activation maps of the last convolutional layer due to its high-level semantics. Previous methods produced heatmaps for intermediate layers that were noisy and non-class specific, reinforcing this focus. However, not only the last convolutional layer impacts a model's output; the effects of deep and intermediate layers also require analysis. In deep-layered models, gradients can become noisy and discontinuous, calling into question their quality as weighting components. Relevance-weighted Class Activation Mapping (Relevance-CAM) addresses this issue by using Layer-wise Relevance Propagation to obtain the weighting components. This method ensures that the explanation map remains faithful and robust against the shattered gradient problem, a common issue in gradient-based CAM methods that results in noisy saliency maps for intermediate layers. Consequently, Relevance-CAM provides a more accurate explanation of a model by analyzing both the intermediate layers and the last convolutional layer\cite{relevancecam}.

Softmax Gradient Layer-wise Relevance Propagation (SGLRP) leverages the gradient of the softmax function to backpropagate the relevance of the output probability to the input image. By using the softmax gradient as the initial relevance from the output layer, the relevance values directly corresponding to the probability of the object belonging to a particular class are propagated. This approach is more natural for eliminating the relevance of non-target classes compared to LRP, which simply disregards other classes, and CLRP, which applies an arbitrarily fixed penalty\cite{slrp}.

FG-CAM (Fine-Grained CAM) extends CAM-based methods to produce detailed and highly accurate explanations. It utilizes the relationship between two adjacent layers of feature maps with different resolutions to incrementally enhance the resolution of explanations. This process identifies and retains contributing pixels while filtering out non-contributing ones. FG-CAM addresses the limitations of traditional CAM methods without altering their fundamental characteristics, resulting in fine-grained explanations that are more faithful than those produced by LRP and its variants. Experimental results demonstrate that FG-CAM's performance remains consistent regardless of explanation resolution. It significantly outperforms existing CAM-based methods in both shallow and intermediate layers, and surpasses LRP and its variants notably at the input layer\cite{fgcam}.

\section{Proposed methodology}
\subsection{Intuition}
Grad-CAM uses the gradients of the target class flowing into the final convolutional layer to produce a coarse localization map of the important regions in the image. It provides a high-level, class-discriminative visualization, highlighting regions in the input that are most relevant for the predicted class. However, it can be coarse and sometimes less precise in highlighting fine details.

LRP redistributes the prediction score backward through the network layers by following the contribution of each neuron, based on a set of propagation rules. It offers a fine-grained and theoretically grounded method to understand which parts of the input contribute most to the output, providing pixel-level explanations. While precise, LRP can sometimes be less intuitive to interpret compared to Grad-CAM’s heatmaps.

The intuition behind combining these methods is to create a more detailed and accurate interpretability framework by merging their complementary strengths.

\begin{enumerate}
    \item Grad-CAM’s Localization + LRP’s Precision: Grad-CAM gives a broader, more general area of importance, while LRP provides exact pixel-wise contributions. Combining them can result in a visualization that is both easy to understand and precise.
    \item Layer-wise Insight: Grad-CAM primarily focuses on the final convolutional layer, whereas LRP can provide insights at different layers of the network. Combining these methods can give a multi-level perspective on the decision process.
    \item Cross-verification: Both of these methods work on different principles. Using both methods together can validate the importance of features. If both methods highlight the same regions, it increases confidence in the interpretation.  
\end{enumerate}

\subsection{Methodology}
\begin{figure}
    \centering
    \includegraphics[width=1\linewidth]{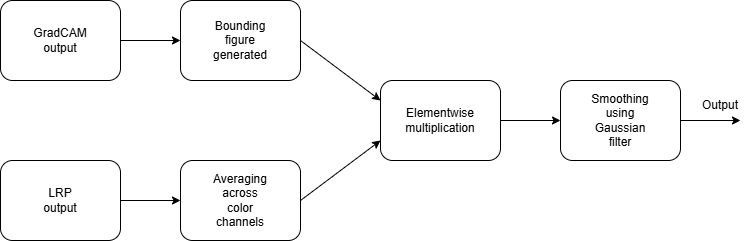}
    \caption{An illustration of the proposed method. First, the GradCAM and LRP output is obtained for the given input image and class. The GradCAM output is processed to remove so that values below a certain threshold become zero. LRP output is averaged across the three color channels. These processed outputs are then multiplied elementwise. Finally, the product is smoothed using a Gaussian filter.}
    \label{fig:architecture}
\end{figure}
\begin{figure*}[!t]
    \centering
    \includegraphics[width=1\linewidth]{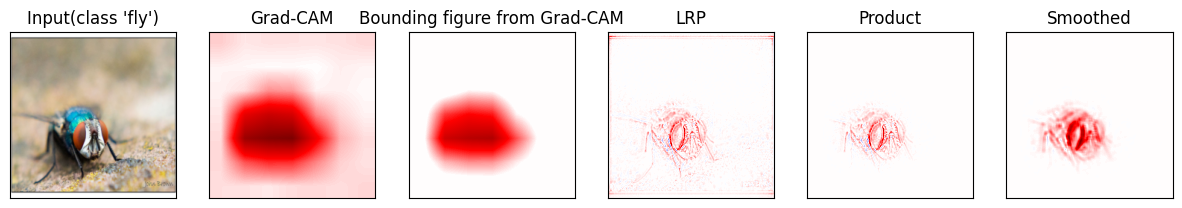}
    \caption{An example of the working of the proposed method.}
    \label{fig:example}
\end{figure*}
The proposed method is illustrated in Fig. \ref{fig:architecture}. First, the Grad-CAM and LRP output is calculated for the given input image and class. An example is given in Fig. \ref{fig:example}.
For the Grad-CAM output, we have used the Grad-CAM++ method. Grad-CAM++\cite{GradCAM++} uses a weighted combination of the positive partial derivatives of the last convolutional layer feature maps with respect to a specific class score as weights to produce a visual explanation. Subjective and objective evaluations on standard datasets have shown that Grad-CAM++ indeed provides better visual explanations for a given CNN architecture when compared to Grad-CAM. We have used pytorch-grad-cam library for generating the Grad-CAM++ output\cite{pytorchcam}.

Based on this output, a bounding figure is created. The Grad-CAM explanation generally has a small positive value even for parts of the image which do not influence the decision. These values need to be equalized to zero to produce a tight bounding figure. The process involves subtracting a user-defined threshold value from all pixel values in the Grad-CAM explanation. The resulting values are then thresholded above 0 to eliminate any negative values. Finally, the values are normalized in the range of 0 to 1 using min-max normalization. 

For the LRP output, we used the emerging strategy of a composite application of multiple purposed decomposition rules. It has been shown that this approach does not only yield measurably more representative attribution maps, but also provides a solution against gradient shattering affecting previous approaches, and improves properties related to object localization and class discrimination via attribution. Common among these works is the utilization of LRPε with ε ≪ 1 (or just LRPz) to decompose fully connected layers close to the model output, followed by an application of LRPαβ to the underlying convolutional layers\cite{complrp}. We have used LRPz for the fully connected layers and LRPαβ to the underlying convolutional layers. 

We have used the python Captum library to generate the LRP explanation. Captum is an open source, extensible library for model interpretability built on PyTorch \cite{captum}. The explanation generated is averaged across the three color channels. This makes it easier to interpret visually.

These two processed explanations are then multiplied element-wise. In the processed Grad-CAM output, the pixel values outside the bounding figure have a value of zero. This ensures that the values for pixels outside the bounding figure become zero in the product even if the LRP explanation has positive values for those pixels. Any noise in the LRP explanation is automatically removed.

Finally, the product is smoothed using a Gaussian filter. By applying Gaussian smoothing, the explanation can become more coherent, emphasizing broader regions that are important rather than isolated pixels. This helps in identifying significant patterns or regions in the input that are truly influential in the model’s prediction. Smoother relevance maps are generally easier to interpret. Gaussian smoothing can make the visual representation of the results more human-friendly, allowing for better understanding and communication of which parts of the input are relevant. Gaussian smoothing can improve the robustness of explanations to small perturbations or variations in the input data. This stability is crucial for the reliability of explanations, especially when the input data is slightly altered.

\section{Benchmarking}
\begin{figure*}[!t]
    \centering
    \includegraphics[width=1\linewidth]{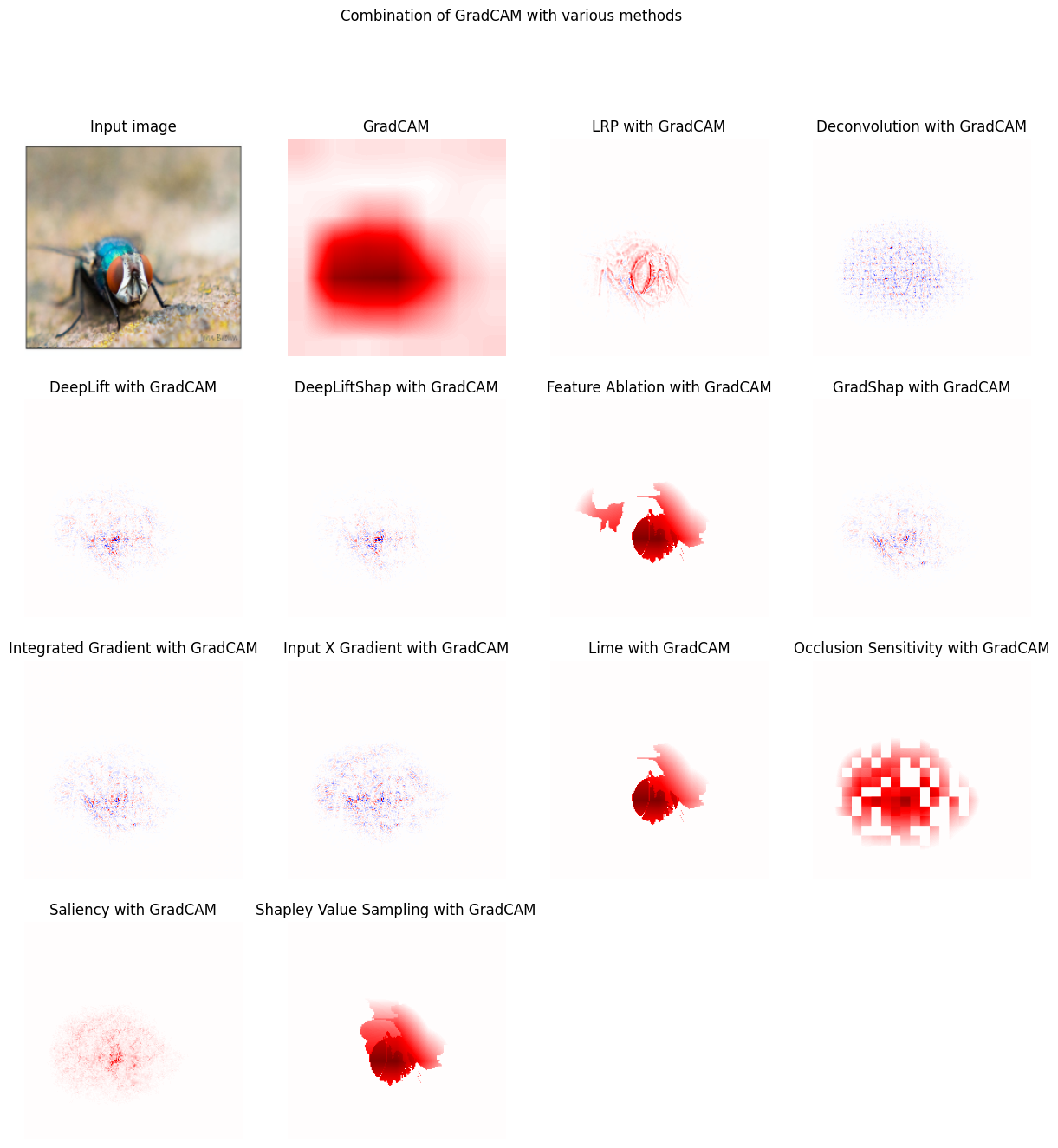}
    \caption{The results of the combination of GradCAM with LRP and other visualization methods. Among these, only the combination of GradCAM with LRP produced comprehensible results, while the other combinations yielded unclear outcomes.}
    \label{fig:combinations}
\end{figure*}
\subsection{Qualitative analysis}
\subsubsection{Comparisions with other combinations}
We combined GradCAM with other explanation methods for CNN-based models and compared the results to that with LRP. The results are shown in fig \ref{fig:combinations}. The input image belongs to the imagenet class 308('fly'). The explanations were generated for VGG-16 model trained on the Imagenet dataset. Captum library was used to generate the results for these methods as well\cite{captum}. In the case of Feature ablation, Lime, Occlusion sensitivity and Shapley value sampling, the pixels were grouped using image segmentation. The importance was then calculated for each group of pixels. The bounding figure generated from the GradCAM explanation was used to remove noises from the explanations produced by the methods.
It was observed that only Grad-CAM with LRP gives comprehensible results. The outline of the fly is clear here. The outline can been seen in Deeplift, DeepliftShap, Feature Ablation, Lime, and Shapley Value Sampling with some difficulty.
We combined GradCAM with various explanation methods for CNN-based models and compared these results with those from LRP. The outcomes are illustrated in fig \ref{fig:combinations}. The input image corresponds to ImageNet class 308 ('fly'). The explanations were generated for a VGG-16 model trained on the ImageNet dataset, using the Captum library to produce these visualizations\cite{captum}.

For methods such as Feature Ablation, LIME, Occlusion Sensitivity, and Shapley Value Sampling, pixels were grouped using image segmentation, and the importance was calculated for each pixel group. We used the bounding figure generated from the GradCAM explanation to reduce noise in the explanations from each method.

We observed that only the combination of GradCAM with LRP yielded comprehensible results, clearly outlining the fly. While the outline is somewhat discernible in the explanations from DeepLIFT, DeepLIFT Shap, Feature Ablation, LIME, and Shapley Value Sampling, it is less clear than in the GradCAM with LRP combination.

\subsubsection{Comparisions with GradCAM and LRP}
\begin{figure*}[!ht]
    \centering
    \includegraphics[width=3.5in]{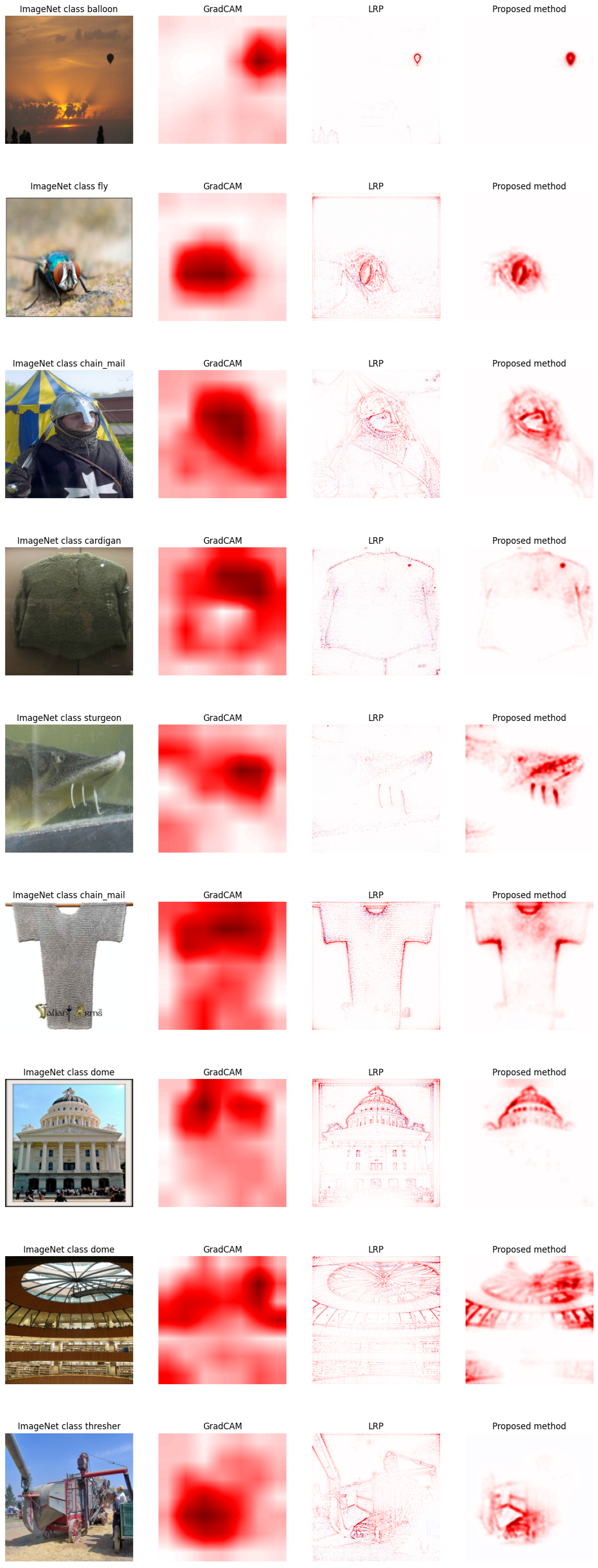}
    \caption{Examples of explanations generated for inputs of various classes on a VGG-16 model trained on imagenet dataset.}
    \label{fig:examples1}
\end{figure*}
\begin{figure*}[!ht]
    \centering
    \includegraphics[width=3.5in]{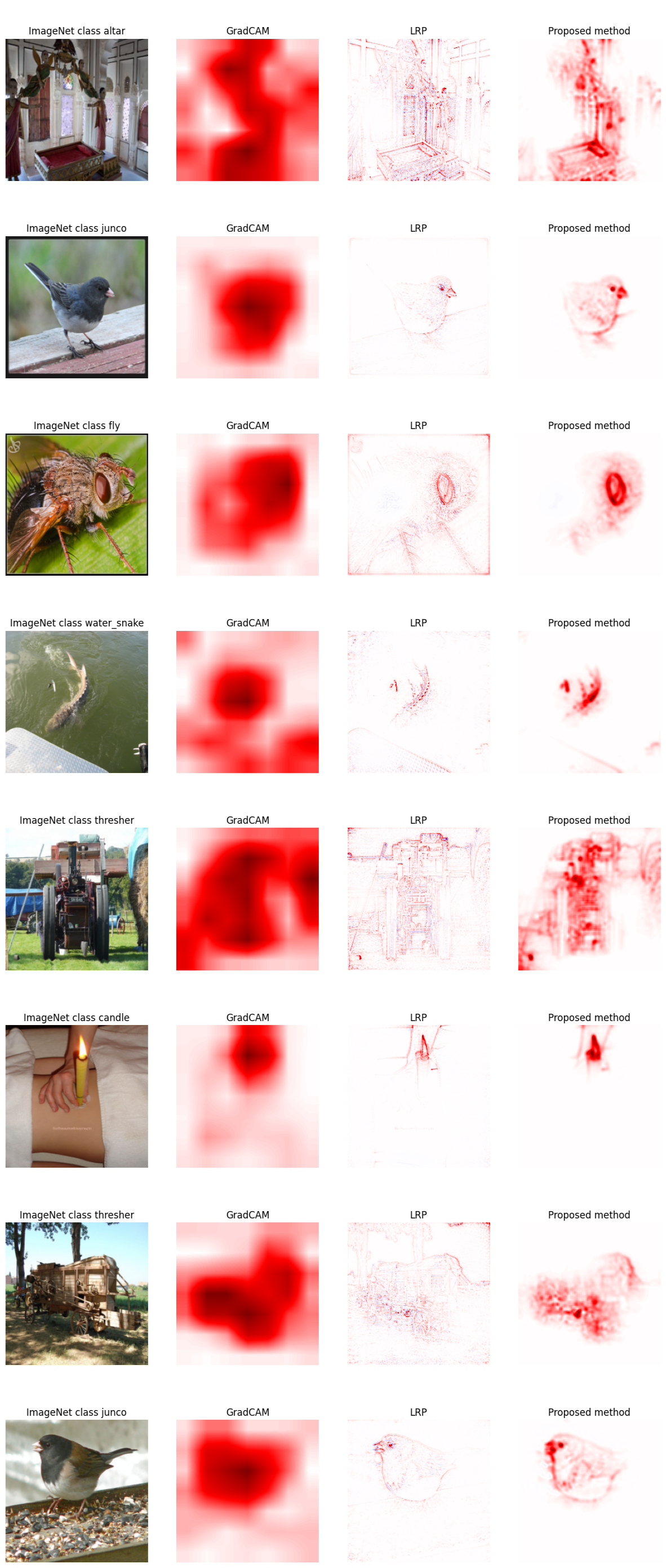}
    \caption{Further examples of explanations generated for inputs of various classes on a VGG-16 model trained on imagenet dataset.}
    \label{fig:examples2}
\end{figure*}
We generated explanations for input images of different classes. The results are shown in Fig. \ref{fig:example} and Fig.\ref{fig:examples2}. The model used was VGG-16 trained on the imagenet dataset. The first column displays the input image with its corresponding ImageNet class indicated above it. The second column shows the Grad-CAM explanation for the input image, while the third column presents the LRP explanation. The final column contains the explanation produced by our proposed combination of Grad-CAM and LRP methods.

Grad-CAM provides relatively coarse localization maps, which do not accurately pinpoint the precise regions of interest in the input image. The heatmaps are spread over broader areas, making it challenging to identify exact features that contribute to the prediction.

LRP provides detailed, pixel-level attributions for each input feature. This granular approach allows users to see exactly which parts of the input contribute to the model's decision. However, it suffers from poor localization. Some of the pixels in the non-relevant regions have positive values even though they should not contribute to the model decision.

The proposed method also provides detailed, pixel-level attributions for each input feature. However, it doesn't suffer from lack of localization like LRP. Due to less noise, the visualization is more clear and intuitive. 

\subsection{Quantitative analysis}
\subsubsection{Infidelity}
\begin{figure}
    \centering
    \includegraphics[width=1\linewidth]{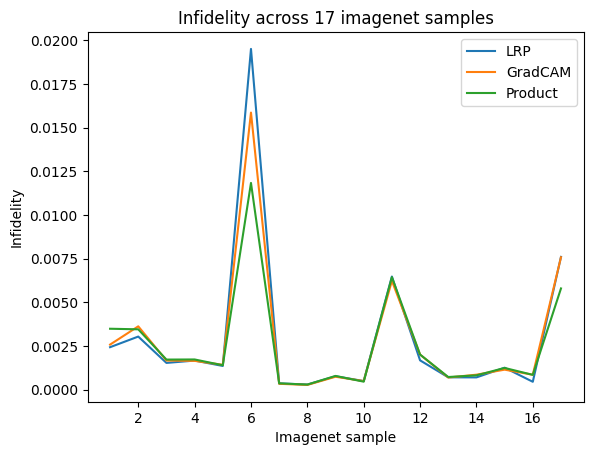}
    \caption{Infidelity values for each method for a sample of images}
    \label{fig:infidelity}
\end{figure}

A natural notion of the goodness of an explanation is to quantify the degree to which it captures how the predictor function itself changes in response to significant perturbations.  Infidelity measure is defined as the expected difference between the two terms: (a) the dot product of the input perturbation to the explanation and (b) the output perturbation (i.e., the difference in function values after significant perturbations on the input)\cite{infidelity}. Lesser the infidelity value, better is the explanation. 
We used the captum library to measure the infidelity for the proposed method across some images sampled from the imagenet dataset. The results are shown in Fig. \ref{fig:infidelity}.
Infidelity values vary a lot even for the same input image. From the graph, it can be observed that infidelity values vary across the sample of images. No explanation method gives the lowest infidelity score across the sample of images. The infidelity values for the proposed method though remains near that of Grad-CAM and LRP.

\subsubsection{Other measures}
We used the Quantus Python library for furthur quantitative analysis of the proposed method. We compared the proposed method on five different areas:
\begin{enumerate}
    \item Faithfulness
    \item Robustness
    \item Complexity
    \item Localization
    \item Randomization
\end{enumerate}

For faithfulness, we used the metric Faithfulness Correlation. The feature importance scores from g should correspond to the important features of x for f; as such, when we set particular features xs to a baseline value xs, the change in predictor’s output should be proportional to the sum of attribution scores of features in xs. Faithfulness Correlation is the correlation between the sum of the attributions of xs and the difference in output when setting those features to a reference baseline\cite{featureCorrelation}.
For robustness, we used the metric Average Sensitivity. We aim to ensure that if inputs are close to each other and their model outputs are similar, then their explanations should also be similar. Assuming the function \( f \) is differentiable, we want the explanation function \( g \) to exhibit low sensitivity around a point of interest \( x \), which implies that \( g \) should be locally smooth\cite{featureCorrelation}.
For complexity, we used the metric Sparseness. Sparseness is measured using the Gini Index applied to the vector of absolute attribution values. The test requires that features genuinely predictive of the output \( F(x) \) have significant contributions, while irrelevant or weakly relevant features should have minimal contributions\cite{sparseness}.
For localization, we used the metric Relevance Rank Accuracy. The Relevance Rank Accuracy evaluates the proportion of high-intensity relevances within the ground truth mask (GT). Given \( P_{top-k} \), the set of pixels sorted by their relevance in decreasing order up to the k-th pixel, the rank accuracy is calculated as: rank accuracy = \( \frac{|P_{top-k} \cap GT|}{|GT|} \). High scores are preferred, indicating that the pixels with the highest positive attribution scores are located within the bounding box of the targeted object\cite{localisation}.

For randomisation, we used the metric Random Logit. The Random Logit Metric measures the distance between the original explanation and the explanation for a randomly chosen non-target class\cite{randomisation}.

\begin{table*}
\caption{Results of Proposed method, Grad-CAM and LRP on the five metrics
\label{tab:metrics}}
    \centering
    \begin{tabular}{|c|ccccc|}
        \hline
          & Robustness & Faithfulness & Localisation & Complexity & Randomisation \\
        \hline
        Proposed method & 1.1653 & 0.03734 & 0.5598 & 0.8355 & 0.8062 \\
        \hline
        GradCAM & 0.2158 & 0.03482 & 0.7459 & 0.4219 & 1.0000 \\
        \hline
        LRP & 1.3243 & 0.08190 & 0.5184 & 0.6286 & 0.6354 \\
        \hline

    \end{tabular}
\end{table*}

\begin{figure}
    \centering
    \includegraphics[width=1\linewidth]{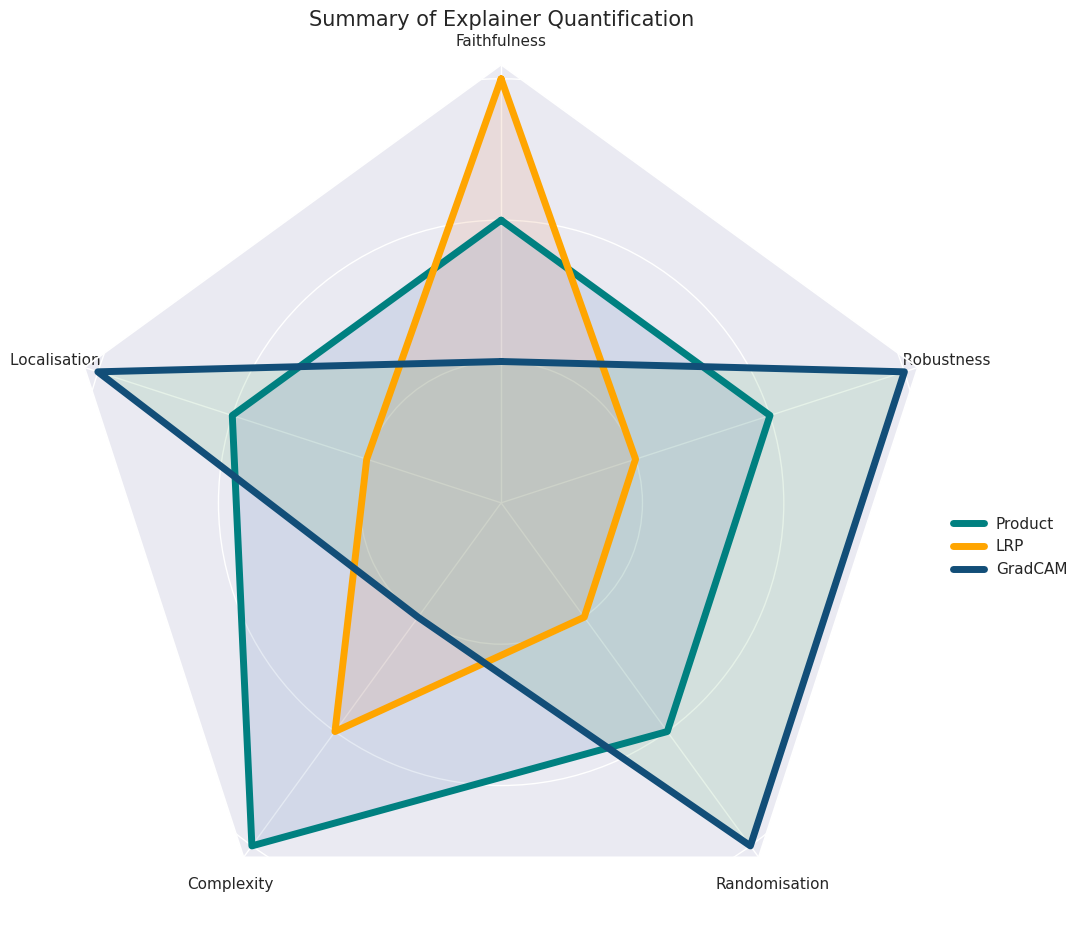}
    \caption{A radar chart illustrating the performance of methods across each metric. A position closer to the outer edge indicates a better performance on that particular metric and vice versa.}
    \label{fig:radar}
\end{figure}

The results are shown in Table \ref{tab:metrics} and summarised in Fig. \ref{fig:radar}. For Robustness, Localisation and Randomisation, Grad-CAM is the best, LRP is the worst and our proposed method lies between them.
For Faithfulness, LRP is the best, Grad-CAM is worst and our proposed method lies between them.
For complexity, our proposed method is the best. This reflects the main strength of our proposed method: sparseness of explanation. Only those pixels are included in the explanation which are present in both Grad-CAM and LRP explanations. This cross-verification ensures that only the most relevant pixels are included in the explanation.

\section{Future Work}
\begin{enumerate}
    \item \textbf{Exploration of Different Architectures:} Investigate how our method performs with different CNN architectures beyond VGG-16. Exploring architectures like ResNet, Inception, or Transformer-based models could provide insights into the generalizability of our approach across diverse network structures.
    \item \textbf{Quantitative Metrics Development:} Develop additional quantitative metrics to assess the interpretability of visual explanations objectively. 
    \item \textbf{Integration with Other Interpretability Techniques:} Explore combinations with other interpretability techniques beyond GradCAM and LRP. Techniques such as Integrated Gradients, SmoothGrad, attention mechanisms, Layer-wise Relevance Propagation (LRP), DeepLIFT, and Occlusion Sensitivity could offer complementary insights and further improve the robustness and clarity of visual explanations.
    \item \textbf{User Studies and Feedback:} Conduct user studies to gather feedback from domain experts and non-experts on the interpretability and usability of our method. Understanding user preferences and requirements could guide further refinement and optimization of the technique for practical applications.
\end{enumerate}

\section{Conclusion}
In this paper, we have proposed a novel interpretability technique that enhances the explainability of CNN-based models by combining GradCAM and LRP. The method capitalizes on the strengths of both techniques: GradCAM's ability to generate broad, class-discriminative localization maps, and LRP's detailed, pixel-level relevance attributions. By processing the GradCAM output to remove noise, performing elementwise multiplication with the LRP output, and applying a Gaussian blur, our approach produces visual explanations that are both clearer and more precise.

Our qualitative analysis, demonstrated through various input images, shows that the combined GradCAM+LRP method offers superior interpretability compared to using GradCAM or LRP alone. The hybrid approach effectively highlights relevant regions in the input image, providing a more balanced and understandable visual representation. This is particularly important for high-stakes applications where understanding the decision-making process of models is crucial.

We assessed the proposed method using key metrics including Faithfulness, Robustness, Complexity, Localization, and Randomization. Our results show that the GradCAM+LRP combination excels in Complexity and surpasses at least one of the individual methods in the other metrics. This indicates that combining GradCAM with LRP can markedly enhance the interpretability and reliability of CNN models.

Overall, this research contributes to the advancement of explainable AI by presenting a method that enhances the clarity and reliability of visual explanations for CNN predictions. The approach promises to foster greater trust and transparency in AI models, which is essential for their broader adoption in critical domains.



 

\bibliographystyle{IEEEtran}
\bibliography{ref}

\newpage

\section{Biography Section}
 

\bf{}\vspace{-33pt}
\begin{IEEEbiography}[{\includegraphics[width=1in,height=1.25in,clip,keepaspectratio]{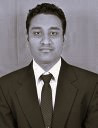}}]{Vaibhav Dhore}
Vaibhav D. Dhore, an accomplished academician, researcher, and educator, holds an M. Tech in Computer Engineering from Malaviya National Institute of Technology, Jaipur, and a B.E. in Computer Science \& Engineering from Government College of Engineering, Amravati.

With over four years of experience in academia, Vaibhav has significantly contributed to the academic community through his teaching, research, and participation in various Faculty Development Programs, including STTP’s\ Summer School-Winter School, where he actively engaged as a participant on three occasions. His commitment to scholarly pursuits is reflected in his substantial research output, with three publications in reputable journals and conferences.

Vaibhav's research interests encompass Machine Learning, High-Performance Computing, and Big Data Analytics. His work aims to address contemporary challenges and push the boundaries of knowledge in these cutting-edge domains.

As an esteemed member of the academic community, Vaibhav D. Dhore continues to inspire and empower students and fellow researchers alike. Through his dedication to advancing the frontiers of computer science and his passion for discovery, Vaibhav serves as a beacon of inspiration for aspiring researchers and scholars worldwide.
\end{IEEEbiography}

\vspace{11pt}

\vspace{-33pt}
\begin{IEEEbiographynophoto}{Achintya Bhat}
is a undergraduate student at Veermata Jijabai Technological Institute studying Information Technology. He will be graduating in 2024.
\end{IEEEbiographynophoto}

\begin{IEEEbiographynophoto}{Viraj Nerlekar}
is a undergraduate student at Veermata Jijabai Technological Institute studying Information Technology. He will be graduating in 2024.
\end{IEEEbiographynophoto}

\begin{IEEEbiographynophoto}{Kashyap Chavhan}
is a undergraduate student at Veermata Jijabai Technological Institute studying Information Technology. He will be graduating in 2024.
\end{IEEEbiographynophoto}

\begin{IEEEbiographynophoto}{Aniket Umare}
is a undergraduate student at Veermata Jijabai Technological Institute studying Information Technology. He will be graduating in 2024.
\end{IEEEbiographynophoto}
\vspace{11pt}


\vfill

\end{document}